\documentclass[10pt,journal,onecolumn]{IEEEtran}
\usepackage{multirow}
\usepackage{algorithmic}
\usepackage{algorithm}
\usepackage{color}
\usepackage{graphicx}
\usepackage{subfig}
\usepackage{amsmath}

\usepackage[T1]{fontenc}
\usepackage[utf8]{inputenc}
\usepackage{authblk}

\author[1]{Ni Zhuang}
\author[1*]{Yan Yan\thanks{*Corresponding author.}}
\author[2]{Si Chen}
\author[1]{Hanzi Wang}
\affil[1]{School of Information Science and Engineering, Xiamen University, Xiamen 361005, China}
\affil[2]{School of Computer and Information Engineering, Xiamen University of Technology, Xiamen 361024, China}
\affil[ ]{Email: ni.zhuang@foxmail.com, \{yanyan, hanzi.wang\}@xmu.edu.cn, chensi@xmut.edu.cn}
\renewcommand{\footnoterule}{%
	\kern 0.1pt
	\hrule width 90pt height 0.02pt
	\kern 2pt 
}
\date{} 

\begin{document}

\title{Multi-task Learning of Cascaded CNN for \\ Facial Attribute Classification}


\maketitle

\begin{abstract}
Recently, facial attribute classification (FAC) has attracted significant attention in the computer vision community. Great progress has been made along with the availability of challenging FAC datasets. However, conventional FAC methods usually firstly pre-process the input images (i.e., perform face detection and alignment) and then predict facial attributes. These methods ignore the inherent dependencies among these tasks (i.e., face detection, facial landmark localization and FAC). 
Moreover, some methods using convolutional neural network are trained based on the fixed loss weights without considering the differences between facial attributes.
In order to address the above problems, we propose a novel multi-task learning of cascaded convolutional neural network method, termed MCFA, for predicting multiple facial attributes simultaneously. 
Specifically, the proposed method takes advantage of three cascaded sub-networks (i.e., S\_Net, M\_Net and L\_Net corresponding to the neural networks under different scales) to jointly train multiple tasks in a coarse-to-fine manner, which can achieve end-to-end optimization.
Furthermore, the proposed method automatically assigns the loss weight to each facial attribute based on a novel dynamic weighting scheme, thus making the proposed method concentrate on predicting the more difficult facial attributes.
Experimental results show that the proposed method outperforms several state-of-the-art FAC methods on the challenging CelebA and LFWA datasets.

\end{abstract}


%

\section{Introduction}
During the past few years, facial attribute classification (FAC) has attracted increasing attention in computer vision and pattern recognition. The goal of FAC is to predict the attributes of a given facial image, such as attraction, smile and gender, as shown in Fig.~1. FAC has widespread and practical applications, including face verification \cite{sun2015deeply3, kumar2009attribute4}, image search \cite{kumar2011describable8} and image retrieval \cite{Siddiquie2011Image}. However, it still remains a challenging task in practice due to the wide variability of facial appearances in viewpoint, illumination, expression, etc.\\
\indent Recently, most methods of FAC take advantage of convolutional neural network (CNN) for predicting facial attributes, due to the outstanding performance of CNN. Generally speaking, the CNN based FAC methods can be roughly divided into two categories: single-label learning based methods \cite{zhang2014panda11,liu2015deep12,zhong2016leveraging14,kang2015face13} and multi-label learning based methods \cite{jiang2016face,Wang2017Deep}. The single-label learning based methods first extract the features of facial images via CNN and then use the support vector machine (SVM) for predicting facial attributes individually. 
In contrast, multi-label learning based methods predict multiple facial attributes simultaneously. 
For example, Ehrlich et al. \cite{ehrlich2016facial} propose a novel multi-task restricted boltzmann machine (MT-RBM) method for FAC.  
Rudd et al. \cite{rudd2016moon} use multi-label learning for FAC by introducing a novel mixed object optimization network (MOON).
Zhuang et al. \cite{Zhuang2018Multi} propose a multi-label learning based deep transfer neural network method for FAC, which effectively addresses the problem of facial attributes with unlabelled information.
Typically, the above methods usually first pre-process the input images (i.e., perform face detection and alignment) and then predict facial attributes. 
In other words, these tasks (i.e., face detection, facial landmark localization and FAC) are independently trained, and thus the inherent dependencies among these tasks are ignored. 
Furthermore, for the loss function in these methods, the loss weights corresponding to different facial attributes are often fixed to be equal. As a result, different facial attributes are equally treated and trained. Therefore, the performance of FAC may not be optimal since the differences between facial attributes are not effectively exploited.\\
\begin{figure}[!t]
\centering
\includegraphics[width=3.0in]{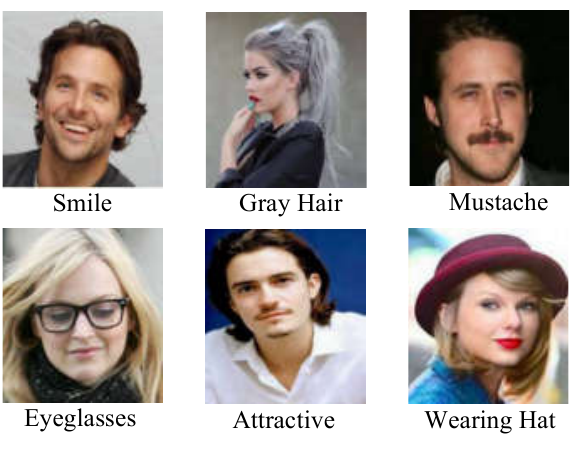}
\caption{Exemplars of different facial attributes.}
\label{fig:1}
\end{figure}
\indent Driven by the above problems, in this paper we propose a novel multi-task learning of cascaded CNN method, termed MCFA, for FAC. 
Three different but related tasks (i.e., face detection, facial landmark localization and FAC) are simultaneously trained by taking advantage of multi-task learning. Notice that FAC is our target task and the other two tasks (i.e., face detection and facial landmark localization) are the auxiliary tasks. We aim to exploit the inherent dependencies between the target task and auxiliary tasks to improve the performance of FAC.
More specifically, the proposed MCFA method jointly trains three cascaded sub-networks (i.e., S\_Net, M\_Net and L\_Net) under three different scales in a unified framework for FAC. The whole framework performs multi-task learning in a coarse-to-fine manner.
In addition, a novel dynamic weighting scheme is proposed for automatically assigning the loss weights corresponding to the facial attributes. In this manner, the CNN training stage focuses on classifying the more difficult facial attributes, thus leading to performance improvements.\\
\indent The major contributions of this paper are summarized as follows:\\
\indent $\bullet$ We propose a novel and effective method for FAC, which takes advantage of multi-task learning to train three related tasks simultaneously. To the best of our knowledge, it is the first work to perform multi-task learning in a unified framework for predicting multiple facial attributes simultaneously. As a result, the performance of FAC can be effectively improved by exploiting the inherent dependencies among different tasks.\\
\indent $\bullet$ Different from the conventional cascaded method \cite{zhang2016mtcnn} which independently trains each stage, the proposed MCFA method jointly trains different sub-networks in a cascaded manner. Therefore, the proposed MCFA method can achieve end-to-end optimization for better performance.\\
\indent $\bullet$ Instead of using the fixed loss weights in the loss function, a dynamic weighting scheme, which automatically and dynamically learns the loss weights, is developed by considering the differences between facial attributes. Thus, the proposed MCFA method concentrates on predicting the more difficult facial attributes.
\section{The Proposed Method}
In this section, we describe the proposed MCFA method in detail, which uses multi-task learning of cascaded CNN, for FAC.
\subsection{Overall Framework}
The overall framework of the proposed MCFA method for FAC is shown in Fig.~2. In this paper, we explicitly extract multi-scale features in a coarse-to-fine manner to boost the performance of FAC. Specifically, the input image is initially resized to three different scales by average pooling to build an image pyramid, which corresponds to the inputs of the following three cascaded sub-networks.\\
\indent \textbf{S\_Net:} This is the first sub-network, which is a fully convolutional network, called S\_Net (Small\_Net), in which the input image is resized to 56$\times$56. S\_Net is used to extract the coarse features of the input image. For training, we employ multi-task learning to jointly train the sub-network for three related tasks. For testing, we also use the threshold control layers (i.e., the `Cls scorel' layers as shown in Fig.~2) to decide which facial proposals from S\_Net contribute to M\_Net. \\
\indent \textbf{M\_Net:} This sub-network is similar to the first sub-network, called M\_Net (Medium\_Net), in which the input image is resized to 112$\times$112. M\_Net is used to extract the fine features of the input image. For training, we jointly train the sub-network for multiple related tasks by multi-task learning. For testing, only the passed facial proposals from M\_Net, which further rejects many false candidates, are fed to L\_Net. \\
\indent \textbf{L\_Net:} This is the third sub-network, called L\_Net (Large\_Net), in which the input image is resized to 224$\times$224. L\_Net is used to extract the subtle features of the input image. For training, the sub-network is learned based on multi-task learning. For testing, only the passed facial proposals are used for performing multiple tasks, where FAC is the target task, while face detection and facial landmark localization are the auxiliary tasks. Therefore, we obtain the final results of FAC from L\_Net. \\
\indent Notice that, different from the method of \cite{zhang2016mtcnn},  which uses three cascaded stages of CNN for joint face detection and alignment in a greedy manner (i.e., previous stages are fixed when training the subsequent stages), the proposed cascaded sub-networks are effectively combined in a unified framework. In this manner, we can easily apply the back propagation algorithm to train the whole framework.\\
\indent Finally, the whole framework combines the losses from the three sub-networks as the final loss.  
\begin{figure*}[!t]
\centering
\centerline{\includegraphics[width=6.9in,height=4.9in]{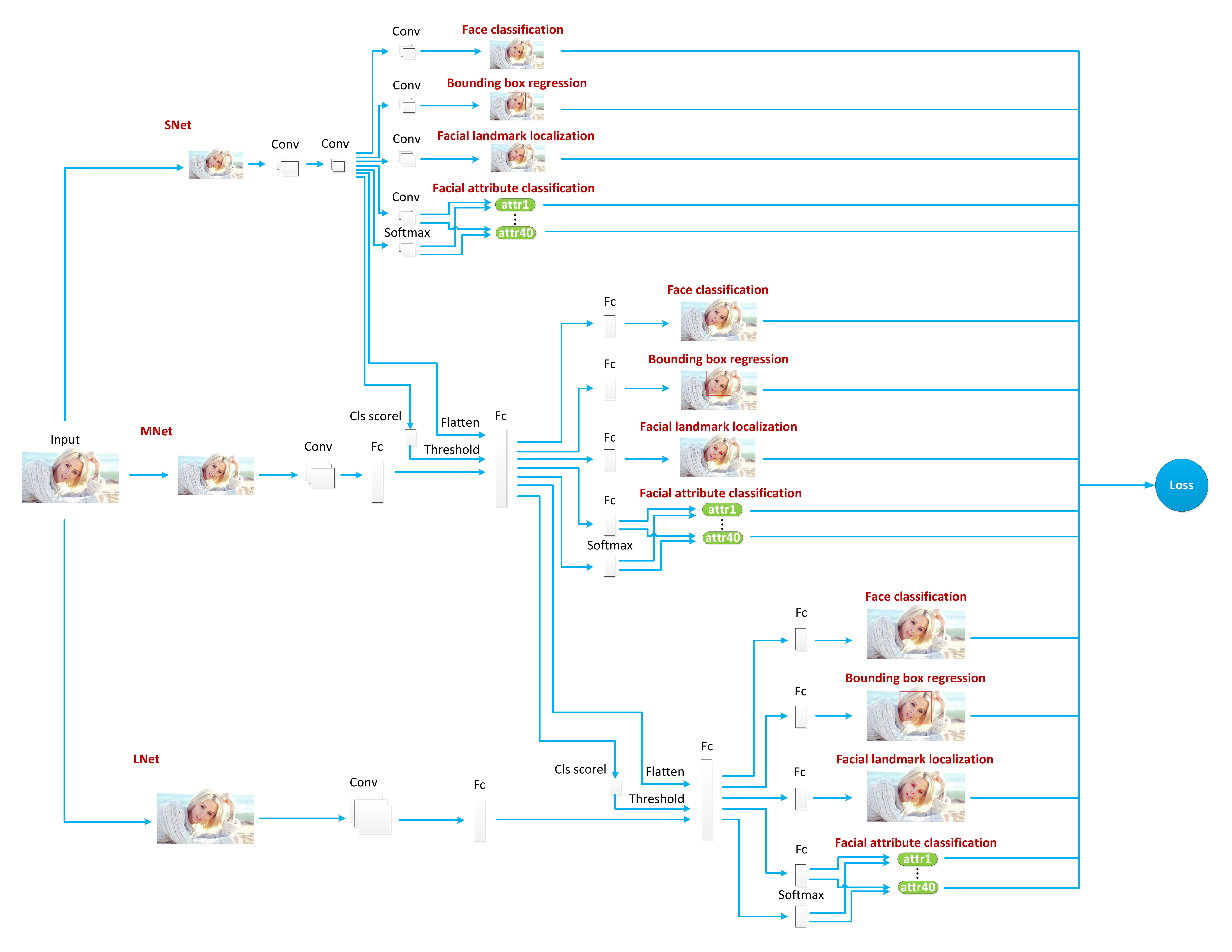}}
\caption{The overall framework of the proposed MCFA method for FAC. The input image is resized to three different scales (i.e., 56$\times$56, 112$\times$112 and 224$\times$224), which correspond to the inputs of the three cascaded sub-networks (i.e., S\_Net, M\_Net and L\_Net).}
\label{fig:2}
\end{figure*}
\subsection{CNN Architecture}
As stated previously, the proposed MCFA method consists of three sub-networks: S\_Net, M\_Net and L\_Net. The detailed architectures of these sub-networks are given as follows:\\
\indent \textbf{S\_Net:} We use the similar architecture of VGG-16 \cite{simonyan2014very29} (from `conv1\_1' to `conv3\_3') for S\_Net in the former convolutional layers. The last shared convolutional layer of S\_Net is of size 1$\times$256$\times$1$\times$1.\\
\indent \textbf{M\_Net:} In the former convolutional layers, we use the similar architecture of VGG-16 \cite{simonyan2014very29} (from `conv1\_1' to `conv4\_3') for M\_Net. Then M\_Net outputs a 1024 dimensional fully connected layer, and it concatenates with a 1$\times$256 fully connected layer, which is flattened from the last shared convolutional layer of S\_Net. Therefore, the last shared fully connected layer of M\_Net is of size 1$\times$1280. \\
\indent \textbf{L\_Net:} Similar to M\_Net, we use the architecture of VGG-16 \cite{simonyan2014very29} (from `conv1\_1' to `conv5\_3') for L\_Net in the former convolutional layers. Then L\_Net outputs a 1024 dimensional fully connected layer, which concatenates with a 1$\times$1280 fully connected layer from M\_Net. \\
\indent Note that, after the last shared layer of each sub-network, we use two layers (i.e., two convolutional layers in S\_Net and two fully connected layers in M\_Net and L\_Net) to learn the more discriminative and compact features for obtaining better performance.
\subsection{Training}
In the proposed method, we jointly train two auxiliary tasks (face detection and facial landmark localization) for the target task (FAC) by taking advantage of multi-task learning. In this way, the inherent dependencies among these tasks are effectively exploited. Note that, the task of face detection includes two sub-tasks: face classification and bounding box regression.
\subsubsection{Face classification}
Face classification is to perform face/non-face classification for a given image. For each sample $\boldsymbol{x}_{i}$, we use the softmax loss:
\begin{equation}
L_{i}^{cls} = y_{i}^{cls}log(p_{i})+(1-y_{i}^{cls})log(1-p_{i}),
\end{equation}
where $p_{i}$ represents the probability given by the network, which indicates the sample $\boldsymbol{x}_{i}$ being a face. And $y_{i}^{cls}\in\lbrace0,1\rbrace$ is the ground-truth label.
\subsubsection{Bounding box regression}
Bounding box regression aims to predict the coordinates of the facial proposal, which is treated as a regression problem. We employ the Euclidean loss for each sample $\boldsymbol{x}_{i}$:
\begin{equation}
L_{i}^{box} = \|\boldsymbol{\hat{y}}_{i}^{box}-\boldsymbol{y}_{i}^{box}\|_{2}^{2},
\end{equation} where $\boldsymbol{\hat{y}}_{i}^{box} \in R^{4}$ represents the coordinate vector (including left, top, height and width) of a regressed facial proposal obtained from the network and $\boldsymbol{y}_{i}^{box}$ denotes the ground-truth coordinate vector.
\subsubsection{Facial landmark localization}
Similar to bounding box regression, facial landmark localization is also a regression problem. We use the Euclidean loss for each sample $\boldsymbol{x}_{i}$ as follows:
\begin{equation}
L_{i}^{landmark} = \|\boldsymbol{\hat{y}}_{i}^{landmark}-\boldsymbol{y}_{i}^{landmark}\|_{2}^{2},
\end{equation}where $\boldsymbol{\hat{y}}_{i}^{landmark} \in R^{2k}$ represents the coordinate vector of the facial landmarks ($k$ is the number of facial landmarks and we use 5 facial landmarks, as done in \cite{zhang2016mtcnn}) obtained from the network and $\boldsymbol{y}_{i}^{landmark}$ denotes the ground-truth coordinate vector.
\subsubsection{Facial attribute classification}
Similar to face classification, we consider FAC as a binary classification problem. The softmax loss with a dynamic weighting scheme for each sample $\boldsymbol{x}_{i}$ is formulated as:
\begin{equation}
L_{i}^{attr} = \boldsymbol{\mu}_{w}^{T} \cdot \boldsymbol{l}_{i}^{cls},
\end{equation} where $\boldsymbol{\mu}_{w}\in R^{d}$ (see Eq.~(6)) denotes the dynamic weight vector corresponding to the $d$ facial attributes and $\boldsymbol{l}_{i}^{cls} = [l_{1}^{cls},l_{2}^{cls},\cdots,l_{d}^{cls}]^{T}$ represents the softmax loss vector corresponding to the $d$ facial attributes for each sample $\boldsymbol{x}_{i}$. 
\subsubsection{Joint loss}
The joint loss consists of the losses of three cascaded sub-networks (i.e., S\_Net, M\_Net and L\_Net). The joint loss of the proposed MCFA method can be written as:
\begin{equation}
L_{joint} = \sum\limits_{i=1}^{N}\sum\limits_{j=1}^{3}(L_{ij}^{cls}+L_{ij}^{box}+L_{ij}^{landmark}+L_{ij}^{attr}),
\end{equation}where $N$ is the number of training examples; $j$ corresponds to the index of three sub-networks. The whole network of the proposed MCFA method can be optimized through the back-propagation algorithm.
\subsubsection{The dynamic weighting scheme}
Inspired by \cite{Yin2017Multi}, we propose to use a dynamic weighting scheme to automatically assign the loss weights to facial attributes.\\
\indent We add a layer (i.e., the convolutional layer in S\_Net and the fully connected layer in both M\_Net and L\_Net) and a softmax layer to the last shared features $\boldsymbol{x}$ of each sub-network for learning the dynamic weights. Let $\boldsymbol{\omega}_{w} \in R^{D\times d}$ and $\boldsymbol{\varepsilon}_{w} \in R^{d}$ respectively represent the weight matrix and the bias vector in the convolutional layer or fully connected layer. Then, the output of the softmax layer is
\begin{equation}
\boldsymbol{\mu}_{w} = softmax(\boldsymbol{\omega}_{w}^{T}\boldsymbol{x} + \boldsymbol{\varepsilon}_{w}),
\end{equation}
where,
\begin{equation}
softmax(\boldsymbol{\alpha}) = [\dfrac{e^{\alpha_{1}}}{\sum\limits_{q=1}^{d}e^{\alpha_{q}}},\dfrac{e^{\alpha_{2}}}{\sum\limits_{q=1}^{d}e^{\alpha_{q}}},\cdots,\dfrac{e^{\alpha_{d}}}{\sum\limits_{q=1}^{d}e^{\alpha_{q}}}]^{T},
\end{equation}
and $\boldsymbol{\mu}_{w} = [\mu_{1},\mu_{2},\cdots,\mu_{d}]^{T}$ represents the $d$-dimensional dynamic weight vector with $\mu_{1}+\mu_{2}+\cdots+\mu_{d}=1.0$.  $\boldsymbol{\alpha} = [\alpha_{1},\alpha_{2},\cdots,\alpha_{d}]^{T}$ represents a $d$-dimensional vector. The $softmax$ function converts the $d$ dynamic weights to positive values that sum to 1.0. \\
\indent According to Eqs.~(5) and (6),  in order to reduce the joint loss, the more difficult facial attributes will be assigned with larger loss weights. Compared with the conventional methods \cite{liu2015deep12, ehrlich2016facial} that use the equal loss weights for all the facial attributes, the proposed dynamic weighting scheme automatically learns the loss weights minimizing the final joint loss. Therefore, the proposed network can focus on the classification of the more difficult attributes by considering the differences between facial attributes. 
\section{Experiments}
In this section, we firstly introduce the two public FAC datasets used for evaluation. Then we discuss the influence of multi-task learning on the final performance. Finally we compare the proposed MCFA method with several state-of-the-art methods.
\subsection{Datasets}
To evaluate the effectiveness of the proposed MCFA method, we employ two challenging FAC datasets: the CelebA \cite{sun2014deep7} and LFWA \cite{huang2007labeled32} datasets. The CelebA dataset is collected in the wild, which contains 202,599 facial images of 10,177 identities, and 40 binary attribute annotations for each facial image. 162,770 images, 19,867 images and 19,962 images are used for training, validation and testing, respectively. The LFWA dataset is another labelled facial attributes dataset in the wild, which has more than 1,680 identities and contains more than 13,000 facial images. The images in this dataset have 73 binary facial attribute annotations.\\
\indent In this paper, we jointly train three tasks (i.e., face detection, facial landmark localization and FAC). Therefore, we use four different kinds of data annotations for training: 1) Non-faces: we randomly crop several patches, whose Intersection-over-Union (IoU) ratios are less than 0.001 to the ground-truth face, from the CelebA dataset; 2) Faces: we detect the faces from the CelebA or the LFWA dataset; 3) Landmarks: we collect five landmarks of the facial images from the CelebA or the LFWA dataset; 4) Facial attributes: we use the images with labelled facial attributes from the CelebA or LFWA datasets. 
\subsection{Influence of multi-task learning}
In this subsection, to show the effectiveness of the proposed MCFA method, we evaluate several variants of the proposed MCFA method, which perform multi-task learning by combining different tasks.\\
\indent Specifically, `MCFA\_FLL\_FAC' represents that we jointly train the tasks of facial landmark localization and FAC. `MCFA\_FD\_FAC' represents that we jointly train the tasks of face detection and FAC. `MCFA\_FAC' represents that we only train the task of FAC. `MCFA' represents that we jointly train three related tasks (i.e., face detecion, facial landmark localization and FAC). \\
\begin{table}[t]
\caption{The detailed settings of the training and testing data on the CelebA and LFWA datasets.}
\center
\scalebox{0.79}{
\begin{tabular}{c|c|c|c|c}
\hline
 &
\multicolumn{2}{c|}{CelebA} &
\multicolumn{2}{|c}{LFWA}\\
\cline{2-5}
Methods  & Training Data & Testing Data  & Training Data & Testing Data \\
\hline
MCFA\_FLL\_FAC & 162,770 &\multirow{4}{*}{19,962} & 13,144 &\multirow{4}{*}{6,571}\\
MCFA\_FD\_FAC & 488,310 & & 19,716&   \\
MCFA\_FAC & 162,770 &  & 6,572 &\\
MCFA & 488,310 & & 26,288 &\\
\hline
\end{tabular}}
\end{table}
\indent The detailed settings of the training and testing data used for different variants of MCFA methods are shown in Table I. The performance obtained by the different variants of the proposed MCFA method on the CelebA dataset is given in Fig.~3.\\ 
\indent As shown in Fig.~3, we can see that `MCFA' achieves better results than the other three variants of the proposed MCFA method (i.e., `MCFA\_FLL\_FAC', `MCFA\_FD\_FAC' and `MCFA\_FAC'). Therefore, by combining FAC and two auxiliary tasks, the performance of FAC can be effectively improved, which demonstrates the effectiveness of multi-task learning of cascaded CNN. In this manner, the relationship among these tasks is fully exploited.
\begin{figure*}[!t]
\centering
\includegraphics[width=6.1in]{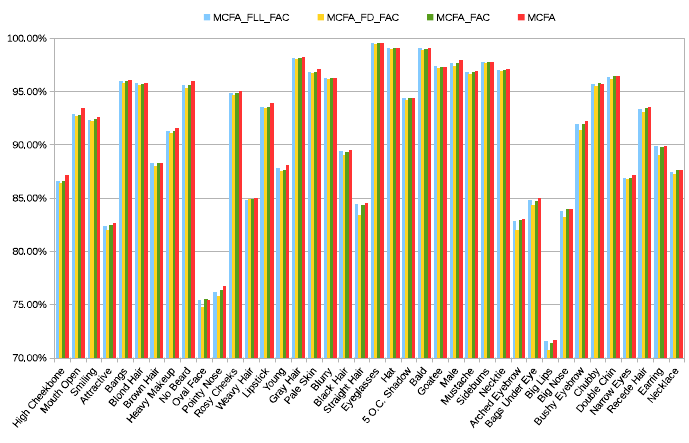}
\caption{Performance comparison between different variants of the proposed MCFA method on the CelebA dataset.}
\label{fig:3}
\end{figure*}
\subsection{Comparison with state-of-the-art methods}
In this subsection, we compare the performance of the proposed MCFA method with several state-of-the-art FAC methods: FaceTracer \cite{kumar2008facetracer33}, two versions of PANDA \cite{zhang2014panda11} (i.e., PANDA-w and PANDA-l), two versions of LNets+ANet \cite{liu2015deep12} (i.e., LNets+ANet (w/o) (without pre-training) and LNets+ANet), and MT-RBM (PCA) \cite{ehrlich2016facial}. In Table II, we show the classification accuracy obtained by the proposed MCFA method and the state-of-the-art methods. For the CelebA dataset, we compare the proposed MCFA method with all the competing methods. For the LFWA method, we compare the proposed MCFA method with all the competing methods except for MT-RBM (PCA) \cite{ehrlich2016facial}.\\ 
\indent As shown in Table II, the proposed MCFA method significantly outperforms most of the competing methods on the CelebA and LFWA datasets. The other competing methods independently train different tasks in a greedy manner, thus the performance of FAC may not be optimal. In contrast, the proposed MCFA method jointly trains these tasks, which makes use of the inherent dependencies among these tasks. Moreover, the proposed method performs multi-task learning in a coarse-to-fine manner, which extracts the hierarchical features for improving the performance. It is worthy pointing out that the proposed MCFA method considers the differences between facial attributes by using a dynamic weighting scheme. Hence, the improvements for some difficult facial attributes are more obvious (such as the `Bangs' and `Wavy Hair' attributes). On the whole, the proposed method achieves superior performance on these two datasets.
\section{Conclusion}
In this paper, we propose a novel multi-task learning of cascaded CNN method, term MCFA, for FAC. The proposed MCFA method can effectively improve the performance of FAC by exploiting the inherent dependencies between the target task (FAC) and auxiliary tasks (face detection and facial landmark localization). Specifically, MCFA  trains three cascaded sub-networks in a unified framework. Moreover, the proposed MCFA method assigns the loss weights to each facial attribute by using a dynamic weighting scheme, which automatically learns the differences between the facial attributes. 
Experimental results on the CelebA and LFWA datasets show that the proposed MCFA method has achieved significant improvements over several other state-of-the-art methods.

\begin{table*}[!t]
\caption{The classification accuracy (\%) comparison between the proposed MCFA and the state-of-the-art methods on the CelebA and LFWA datasets. The best results are boldfaced.}
\center
\scalebox{0.78}{
\begin{tabular}{c|c|c|c|c|c|c|c|c|c|c|c|c|c|c|c|c|c|c|c|c|c|c}
\hline
 &  & \rotatebox{90}{High Cheekbone} & \rotatebox{90}{Mouth Open} & \rotatebox{90}{Smiling} & \rotatebox{90}{Attractive} & \rotatebox{90}{Bangs} & \rotatebox{90}{Blond Hair} & \rotatebox{90}{Brown Hair} & \rotatebox{90}{Heavy Makeup} & \rotatebox{90}{No Beard} & \rotatebox{90}{Oval Face} & \rotatebox{90}{Pointy Nose} & \rotatebox{90}{Rosy Cheeks} & \rotatebox{90}{Wavy Hair} & \rotatebox{90}{Lipstick} & \rotatebox{90}{Young} & \rotatebox{90}{Gray Hair} & \rotatebox{90}{Pale Skin} & \rotatebox{90}{Blurry} & \rotatebox{90}{Black Hair} & \rotatebox{90}{Straight Hair} & \rotatebox{90}{Eyeglasses} \\
\hline
\multirow{9}{*}{CelebA} 
& FaceTracer \cite{kumar2008facetracer33} &84 &87 &89 &78 &88 &80 &60 &85 &90 &64 &68 &84 &73 &89 &80 &90 &83 &81 &70 &63 &98 \\
& PANDA-w  \cite{zhang2014panda11} &80 &82 &89 &77 &89 &81 &69 &84 &87 &62 &65 &81 &76 &88 &77 &88 &84 &77 &74 &67 &94  \\
& PANDA-l  \cite{zhang2014panda11} &86 &93 &92 &81 &92 &93 &77 &90 &93 &65 &71 &87 &77 &93 &84 &94 &91 &86 &85 &69 &98  \\
& ANet  \cite{li2013learning} &85 &85 &92 &79 &94 &86 &74 &87 &91 &65 &67 &85 &79 &91 &81 &93 &89 &83 &77 &70 &96  \\
& LNets+ANet(w/0)  \cite{liu2015deep12} &84 &86 &88 &77 &92 &91 &78 &85 &92 &63 &70 &87 &75 &90 &83 &93 &87 &80 &84 &69 &96   \\
& LNets+ANet  \cite{liu2015deep12} &\textbf{87} &92 &92 &81 &95 &95 &80 &90 &95 &66 &72 &90 &80 &93 &87 &97 &91 &84 &88 &73 &99\\
& MT-RBM (PCA)  \cite{ehrlich2016facial} &83 &82 &88 &76 &88 &91 &83 &85 &90 &73 &73 &94 &72 &89 &81 &97 &96 &95 &76 &80 &96  \\
& MCFA &\textbf{87} &\textbf{93} &\textbf{93} &\textbf{83} &\textbf{96} &\textbf{96} &\textbf{88} &\textbf{92} &\textbf{96} &\textbf{75} &\textbf{77} &\textbf{95} &\textbf{85} &\textbf{94} &\textbf{88} &\textbf{98} &\textbf{97} &\textbf{96} &\textbf{89} &\textbf{85} &\textbf{100}  \\
\hline
\hline
\multirow{9}{*}{LFWA} 
& FaceTracer \cite{kumar2008facetracer33} &77 &77 &78 &71 &72 &88 &62 &88 &69 &66 &74 &70 &62 &87 &80 &78 &70 &73 &76 &67 &90   \\
& PANDA-w  \cite{zhang2014panda11} &75 &74 &77 &70 &79 &87 &65 &86 &63 &64 &68 &64 &63 &83 &76 &77 &64 &70 &78 &68 &84   \\
& PANDA-l  \cite{zhang2014panda11} &86 &78 &89 &81 &84 &94 &74 &93 &75 &72 &76 &73 &75 &93 &82 &81 &\textbf{84} &74 &87 &73 &89   \\
& ANet  \cite{li2013learning} &79 &76 &82 &75 &84 &90 &71 &89 &69 &66 &72 &71 &65 &86 &79 &82 &68 &75 &82 &72 &88   \\
& LNets+ANet(w/0)  \cite{liu2015deep12} &83 &78 &88 &80 &84 &94 &73 &91 &75 &71 &76 &72 &73 &92 &82 &81 &81 &70 &86 &71 &92   \\
& LNets+ANet  \cite{liu2015deep12} &\textbf{88} &\textbf{82} &\textbf{91} &\textbf{83} &88 &\textbf{97} &\textbf{77} &\textbf{95} &\textbf{79} &\textbf{74} &\textbf{80} &78 &76 &\textbf{95} &86 &84 &\textbf{84} &74 &90 &76 &\textbf{95}   \\
& MCFA  &85 &78 &88 &77 &\textbf{89} &\textbf{97} &\textbf{77} &94 &\textbf{79} &\textbf{74} &\textbf{80} &\textbf{85} &\textbf{79} &94 &\textbf{87} &\textbf{88} &82 &\textbf{86} &\textbf{91} &\textbf{77} &91   \\
\hline
\hline
 &  & \rotatebox{90}{Hat} & \rotatebox{90}{5 O.C. Shadow} & \rotatebox{90}{Bald} & \rotatebox{90}{Goatee} & \rotatebox{90}{Male} & \rotatebox{90}{Mustache} & \rotatebox{90}{Sideburns} & \rotatebox{90}{Necktie} & \rotatebox{90}{Arched Eyebrow} & \rotatebox{90}{Bags Under Eye} & \rotatebox{90}{Big Lips} & \rotatebox{90}{Big Nose} & \rotatebox{90}{Bushy Eyebrow} & \rotatebox{90}{Chubby} & \rotatebox{90}{Double Chin} & \rotatebox{90}{Narrow Eyes} & \rotatebox{90}{Recede Hair} & \rotatebox{90}{Earring} & \rotatebox{90}{Necklace} & & \rotatebox{90}{\textbf{Average}}\\
\hline
\multirow{9}{*}{CelebA} 
& FaceTracer \cite{kumar2008facetracer33} &89 &85 &89 &93 &91 &91 &94 &86 &76 &76 &64 &74 &80 &86 &88 &82 &76 &73 &68 & &81   \\
& PANDA-w  \cite{zhang2014panda11} &91 &82 &92 &86 &93 &83 &90 &88 &73 &71 &61 &70 &76 &82 &85 &79 &82 &72 &67 & &79   \\
& PANDA-l  \cite{zhang2014panda11} &96 &88 &96 &93 &97 &93 &93 &91 &78 &79 &67 &75 &86 &86 &88 &84 &85 &78 &67 & &85   \\
& ANet  \cite{li2013learning} &93 &86 &92 &92 &95 &87 &94 &90 &75 &77 &63 &74 &80 &86 &90 &83 &84 &77 &70 & &83   \\
& LNets+ANet(w/0)  \cite{liu2015deep12} &96 &88 &95 &92 &94 &91 &91 &86 &74 &73 &66 &75 &85 &86 &88 &77 &85 &78 &68 & &83   \\
& LNets+ANet  \cite{liu2015deep12} &\textbf{99} &91 &98 &95 &\textbf{98} &95 &96 &93 &79 &79 &68 &78 &90 &91 &92 &81 &89 &82 &71 & &87  \\
& MT-RBM (PCA)  \cite{ehrlich2016facial} &97 &90 &98 &96 &90 &\textbf{97} &96 &94 &77 &81 &69 &81 &88 &95 &\textbf{96} &86 &92 &81 &87 & &87  \\
& MCFA &\textbf{99} &\textbf{94} &\textbf{99} &\textbf{97} &\textbf{98} &\textbf{97} &\textbf{98} &\textbf{97} &\textbf{83} &\textbf{85} &\textbf{72} &\textbf{84} &\textbf{92} &\textbf{96} &\textbf{96} &\textbf{87} &\textbf{94} &\textbf{90} &\textbf{88} & &\textbf{91}  \\
\hline
\hline
\multirow{9}{*}{LFWA} 
& FaceTracer \cite{kumar2008facetracer33} &75 &70 &77 &69 &84 &83 &71 &71 &67 &65 &68 &73 &67 &67 &70 &73 &63 &88 &81 & &74   \\
& PANDA-w  \cite{zhang2014panda11} &78 &64 &82 &65 &86 &77 &68 &70 &63 &63 &64 &71 &63 &65 &64 &68 &61 &85 &79 & &71   \\
& PANDA-l  \cite{zhang2014panda11} &82 &84 &84 &75 &92 &87 &76 &79 &79 &80 &73 &79 &79 &69 &75 &73 &84 &92 &86 & &81   \\
& ANet  \cite{li2013learning} &82 &78 &86 &68 &91 &79 &72 &72 &66 &72 &70 &73 &69 &68 &70 &74 &70 &87 &81 & &76   \\
& LNets+ANet(w/0)  \cite{liu2015deep12} &84 &81 &83 &75 &91 &87 &72 &76 &78 &79 &72 &76 &79 &70 &74 &77 &81 &90 &83 & &79   \\
& LNets+ANet  \cite{liu2015deep12} &88 &\textbf{84} &88 &78 &\textbf{94} &\textbf{92} &77 &79 &\textbf{82} &\textbf{83} &\textbf{75} &\textbf{81} &\textbf{82} &73 &\textbf{78} &\textbf{81} &\textbf{85} &\textbf{94} &88 & &\textbf{84}   \\
& MCFA  &\textbf{91} &75 &\textbf{91} &\textbf{80} &93 &91 &\textbf{78} &\textbf{82} &79 &79 &\textbf{75} &\textbf{81} &76 &\textbf{74} &77 &78 &\textbf{85} &93 &\textbf{89} & &\textbf{84}  \\
\hline
\end{tabular}
}
\end{table*}



\section*{Acknowledgments}
This work was supported by the National Key R\&D Program of China under Grant 2017YFB1302400, by the National Natural Science Foundation of China under Grants 61571379, 61503315, U1605252, and 61472334, by the Natural Science Foundation of Fujian Province of China under Grant 2017J01127 and 2018J01576, and by the Fundamental Research Funds for the Central Universities under Grant 20720170045.



%

\end{document}